# Medical-VLBERT: Medical Visual Language BERT for COVID-19 CT Report Generation With Alternate Learning

Guangyi Liu, *Graduate Student Member, IEEE*, Yinghong Liao, *Graduate Student Member, IEEE*,
Fuyu Wang, Bin Zhang, Lu Zhang, Xiaodan Liang, Xiang Wan, Shaolin Li,
Zhen Li, Shuixing Zhang, and Shuguang Cui, *Fellow, IEEE*

*Abstract*—Medical imaging technologies, including computed tomography (CT) or chest X-Ray (CXR), are largely employed to facilitate the diagnosis of the COVID-19. Since manual report writing is usually too time-consuming, a more intelligent auxiliary medical system that could generate medical reports automatically and immediately is urgently needed. In this article, we propose to use the medical visual language BERT (Medical-VLBERT) model to identify the abnormality on the COVID-19 scans and generate the medical report automatically based on the detected lesion regions. To produce more accurate medical reports and minimize the visual-and-linguistic differences, this model adopts an alternate learning strategy with two procedures that are knowledge pretraining and transferring. To be more precise, the knowledge pretraining procedure is to memorize the knowledge from medical texts, while the transferring procedure is to utilize the acquired knowledge for professional medical sentences generations through observations of medical images. In practice, for automatic medical report generation on the COVID-19 cases, we constructed a dataset of 368 medical findings in Chinese and 1104 chest CT scans from The First Affiliated Hospital of Jinan University, Guangzhou, China, and The Fifth Affiliated Hospital of Sun Yat-sen University, Zhuhai, China. Besides, to alleviate the insufficiency of the COVID-19 training samples, our model was first trained on the large-scale Chinese CX-CHR dataset and then transferred to the COVID-19 CT dataset for further fine-tuning. The experimental results showed that Medical-VLBERT achieved state-of-the-art performances on terminology prediction and report generation with the Chinese COVID-19 CT dataset and the CX-CHR dataset. The Chinese COVID-19 CT dataset is available at https://covid19ct.github.io/.

Manuscript received May 12, 2020; revised December 3, 2020 and April 23, 2021; accepted July 11, 2021. This work was supported in part by the Key Area Research and Development Program of Guangdong Province under Grant 2020B0101350001, in part by the Key Area Research and Development Program of Guangdong Province under Grant 2018B030338001, in part by the National Key Research and Development Program of China under Grant 2018YFB1800800, in part by the Shenzhen Outstanding Talents Training Fund, in part by the Guangdong Research Project under Grant 2017ZT07X152, in part by the NSFC-Youth under Grant 61902335, in part by the Guangdong Regional Joint Fund-Key Projects under Grant 2019B1515120039, in part by the National Natural Science Foundation Fund of China under Grant 61931024, in part by Helixon Biotechnology Company Fund, and in part by the CCF-Tencent Open Fund. *(Guangyi Liu, Yinghong Liao, and Fuyu Wang contributed equally to this work.) (Corresponding authors: Xiaodan Liang; Shuixing Zhang; Shaolin Li; and Zhen Li.)*

Guangyi Liu, Yinghong Liao, Zhen Li, and Shuguang Cui are with the Shenzhen Research Institute of Big Data, The Chinese University of Hong Kong, Shenzhen 518172, China, and also with the Future Network of Intelligence Institute (FNii), The Chinese University of Hong Kong, Shenzhen 518172, China (e-mail: guangyiliu@link.cuhk.edu.cn; yinghongliao@link.cuhk.edu.cn; lizhen@cuhk.edu.cn; shuguangcui@cuhk.edu.cn).

Fuyu Wang is with the School of Computer Science and Engineering, Sun Yat-sen University, Guangzhou 510006, China (e-mail: wangfy8@mail2.sysu.edu.cn).

Bin Zhang and Shuixing Zhang are with the Department of Radiology, The First Affiliated Hospital of Jinan University, Guangzhou 510630, China (e-mail: 1297225541@qq.com; shui7515@126.com).

Lu Zhang is with the Faculty of Medical Science, Jinan University, Guangzhou 510630, China (e-mail: zl2019@stu2019.jnu.edu.cn).

Xiaodan Liang is with the School of Intelligent Systems Engineering, Sun Yat-sen University, Guangzhou 510006, China (e-mail: xdliang328@gmail.com).

Xiang Wan is with the Shenzhen Research Institute of Big Data, The Chinese University of Hong Kong, Shenzhen 518172, China, and also with the Guangdong Provincial Key Laboratory of Big Data Computing, The Chinese University of Hong Kong, Shenzhen 518172, China (e-mail: wanxiang@sribd.cn).

Shaolin Li is with The Fifth Affiliated Hospital of Sun Yat-sen University, Zhuhai 519000, China (e-mail: lishlin5@mail.sysu.edu.cn).

## I. INTRODUCTION

**T**HE medical images obtained from the computed tomography (CT) and the Chest X-Ray (CXR) could manifest the resulted pulmonary lesions, such as multiple ground glass opacity and infiltration. Medical imaging technologies are, therefore, extensively employed in coronavirus detection and contribute tremendously to the quick diagnosis of the COVID-19.

However, most medical imaging systems merely present the projections of lung condition and usually generate scans in large quantities, adding huge burdens to radiologists' workload and severely hinder the rapid diagnosis of the COVID-19. Therefore, a more intelligent system with the capacity of analyzing the lesions in images and writing corresponding medical reports automatically is of great significance to the COVID-19 diagnosis, as shown in Fig. 1.

Within such intelligent system illustrated in Fig. 1 for the COVID-19 diagnosis, medical report generation is the core component, which attempts to draw precise connections between lesion areas in images and their relevant pathological analysis in the text [1], [2]. Due to the naturally imbalanced distributions of normal and abnormal case numbers in training data, previous methods tend to focus more on the normal image patches. With the insufficient attention related to lung injuries, they usually produce undesirable findings and conclusions that contain incorrect terminologies. Thus, the proper import and acquisition of the external medical-specific knowledge are the key to the precise detection of the abnormal terminologies and the production of accurate reports. Nevertheless, in previous methods, the simple application of vanilla recurrent neural network (RNN) [3] and its variants [4] in language generation models without the external medical-specific knowledge hinders the model from fully utilizing the visual-and-linguistic information simultaneously.

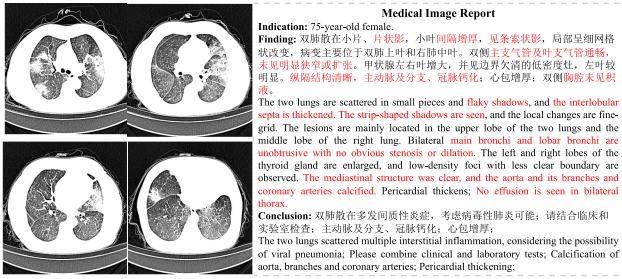

Fig. 1. Example of the COVID-19 medical image report in Chinese. The medical terminologies in findings are marked red. The intelligent system for automatic medical report generation needs to employ these terminologies accurately to offer radiologists a report template with the details of individual lesions, e.g., shape, density, and boundary. With the template, radiologists can not only quickly localize and analyze the lesion areas without the laborious search but also produce a medical report in a short time by a simple modification of the contents.

The emergence of BERT [5] and its variant VL-BERT [6], however, provides a solution to this problem, making it possible for medical report generation system to exploit the external visual-and-linguistic knowledge simultaneously and produce more professional sentences. Equipped with these promising techniques, an efficient AI diagnosis system could be built for the quick diagnosis of the COVID-19 and largely facilitate the physicians' treatment.

However, the superior performances of a deep neural network, e.g., BERT and ResNet [7], usually depend on the scale of the training data. On account of the sudden outbreak of the COVID-19, the inadequate number of training samples concerning COVID-19 becomes the main obstacle for building an AI diagnosis system. Recent studies propose to introduce additional data for model training. Wang *et al.* [8] collected a set of images of viral pneumonia to expand data size. Shi *et al.* [9] added the samples of community-acquired pneumonia to their COVID-19 dataset. However, the model might be misguided by the introduced noisy information of other diseases and fail to capture key features of the COVID-19, resulting in poor performances on report generation.

Therefore, in this article, to efficiently aid radiologists in diagnosis, the Medical Visual Language BERT (Medical-VLBERT) is proposed for automatic medical report generation. To effectively bridge the vision-and-language gap and improve the model performance, we further employ an alternate learning fashion that contains two procedures: knowledge pretraining and knowledge transferring. On the one hand, the pretraining procedure learns to parse and memorize the knowledge contained in medical textbooks. On the other hand, the transferring procedure further utilizes the acquired knowledge to generate medical reports. In practice, both procedures are executed on two major components: 1) *terminology encoder* and 2) *shared language decoder*. Specifically, the *terminology encoder* processes the multimodal features, i.e., medical images and medical reports features, and finds their mutual relation. Then, the *shared language decoder* performs the sentence generation based on the information obtained from *terminology encoder*. For the current study on automatic medical report generation for COVID-19 cases, we are the first to construct a dataset of 368 medical findings in Chinese about 96 patients and 1104 corresponding chest CT scans under the guidance of the Diagnosis and Treatment Protocol for Novel Coronavirus Pneumonia.[1] Instead of combining COVID-19 data with additional information in model training, we adopt a transfer learning strategy. In practice, considering the syntactic similarity shared in the medical report, we first train our model on the large-scale CX-CHR dataset consisting of 45 598 X-ray images and their findings in Chinese. Then, we fine-tune the model on the newly built COVID-19 datasets, using both accurate medical tags and findings on the COVID-19. Note that, even though the screening techniques are different, the medical reports share similar formats so that the transfer can benefit from the pretrained language BERT model.

In summary, our main contributions are fourfold.

1) We present Medical-VLBERT for automatic medical report generation. It is the first model that can produce medical reports for the COVID-19 CT scans to the best of our knowledge.
2) We adopt a transfer learning strategy in model training to alleviate the shortage of the available COVID-19 data. The transferred knowledge provides professional guidance for medical report generation. Our model achieves state-of-the-art performances on both terminology prediction and report generation on the COVID-19 CT dataset.
3) We develop an alternate training strategy for both pretraining and transferring procedures to minimize the discrepancies between the medical scans and the diagnosis texts, as well as to maximize the accuracy of the produced reports.
4) We build a COVID-19 CT dataset that contains 1 104 CT scans and 368 standardized Chinese medical reports by professional radiologists based on the data of 96 patients under the guidance of the Diagnosis and Treatment Protocol for Novel Coronavirus Pneumonia. We have released the COVID-19 CT dataset to the community.[2]

---

[1]The Diagnosis and Treatment Protocol for Novel Coronavirus Pneumonia (Trial Version 8): English version and Chinese version.

[2]Website for COVID-19 CT dataset: https://covid19ct.github.io/

## II. RELATED WORK

### A. Visual Captioning

Visual captioning [10]–[14] aims at generating a descriptive sentence for images or videos. The generated sequence is usually short, describing the main visual information in an image or a video in one sentence. One mainstream for the visual captioning method is the reinforcement learning (RL)-based method, which directly uses the evaluation metrics as the reward function. The agent interacts with the environment by executing actions and receiving rewards, updating policy model parameters via gradient descent. It has gained increasing popularity in sequence generation tasks, such as visual captioning [15], [16] and text summarization [17], [18].

### B. Medical Report Generation

Automatic generation of medical image reports is a crucial application in both academia and industry. The task is similar to image captioning. However, considering the requirements for large amounts of training data and time, the previous RL-based method [1] for visual captioning is not suitable for medical report generation where the data scale is limited , especially in the recent case of the COVID-19. Many researchers have explored this area step by step, especially CXRs' report generation [1], [19], [20]. For example, TieNet [20] classifies the CXRs by using both image features and text embeddings and then transformed the framework into a CXR reporting system.

### C. Language Model Pretraining

Learning language representations from large-scale texts in an unsupervised manner have attracted extensive attention. On the one hand, ELMo [21] extracts context-sensitive word embeddings from a language model and integrates them into task-specific architectures for better feature representations. On the other hand, BERT [5] and OpenAI GPT [22] can generalize to an extensive suite of language understanding tasks without task-specific architecture designed by unsupervised pretraining. However, they are trained in general domain corpora, which is quite different from the medical domain; previous pretrained language models cannot be directly applied to accurate medical report generation.

### D. Transfer Learning for Medical Diagnosis

Considering the significant differences and distributions between medical images, e.g., CT scans and X-ray, and nature images, some measures should be made to embrace the success of deep learning methods for medical diagnosis. Transfer learning [23]–[27] has the advantages of minimizing the domain gaps between different datasets and maximizing the model's capacity of generalization. Inspired by previous progress in medical image segmentation [28], lesion localization [29], and stage diagnosis [30], transfer learning is leverage in our proposed model, which not only transfers the medical knowledge from textbook to the COVID-19 analysis but also addresses the issue of the shortage of the COVID-19 CT data.

### E. AI-Related Research Against COVID-19

In spite of the rapid outbreak of the COVID-19, AI techniques can play an import role in improving and accelerating the COVID-19 diagnosis and treatment. For instance, the 3-D deep learning model [31] can effectively extract infection regions on CT images to increase the accuracy of the COVID-19 detection. Multi-Task Net [32] adopts a transfer learning approach to perform multitask learning for the COVID-19 detection and segmentation on CT and X-ray scans. COVID-19-CT-CXR [33] is a public database of the COVID-19 medical images and texts, but the data are automatically extracted from the COVID-19-relevant articles and remain the issue of poor quality. Therefore, this database may not be an appropriate option for constructing an ideal AI medical diagnosis system. To the best of our knowledge, there does not exist an intelligence system for automatic medical report generation of the COVID-19.

## III. METHODOLOGY

### A. Overview

Our proposed model contains two key procedures: knowledge pretraining and transferring on the COVID-19 CT dataset. For knowledge pretraining, we employ a simple lookup table to generate textbook embeddings $\mathcal{E} = \{e_i\}_{i=1}^{N_e}$, where $e_i \in R^{d_e}$. For the other procedure, we apply DenseNet-121 on CT scans to extract the spatial features $\mathcal{V} = \{v_i\}_{i=1}^{N_v}$, where the pixel-level feature vector is $v_i \in R^{d_v}$. With the given textbook embeddings $\mathcal{E}$ or the visual contexts $\mathcal{V}$, a terminology encoder is further executed to produce terminology-related features $\mathcal{M} = \{m_i\}_{i=1}^{N_m}$, where $m_i \in R^{d_m}$ represents the feature vector of a single terminology. Under the guidance of the terminology-related features $\mathcal{M}$, a language decoder is capable of generating the original textbook sequence $T$ or the report sequence $R$. To bridge cross-modal gaps and achieve the semantic alignment in linguistic and visual domains, two main procedures are performed in an alternate manner via the shared language decoder. The overview of our proposed model is displayed in Fig. 2.

### B. Terminology Encoder

As shown in Fig. 3, the terminology encoder is built to associate the terminology representations with the visual contexts or textbook embeddings. Its architecture is the same in these two procedures except for the data inputs. In this section, the transferring procedure will be elaborated. We use the predefined terminology word embeddings to represent the medical domain knowledge. Then, we align each medical terminology with image regions by their coherent relation. In this manner, for each possible visual context that might imply the disease, the correlated medical terminology can be retrieved and further assisted in medical report generation.

Specifically, the terminology encoder is implemented on VL-BERT [6]. As a variant of BERT [5], VL-BERT encodes the input information on multilayer bidirectional transformer [34] and explores the hidden correlation within input elements. In the architecture of the transformer, each input element jointly interacts with other elements by learning adaptive weights. The learned weight distributions are then transferred to the next layer. By introducing the visual inputs, VL-BERT has the ability of accommodating the linguistic and visual contents simultaneously. Therefore, VL-BERT is used to process multimodal input pairs in our task. In the model of VL-BERT, linguistic and visual contents are processed by the attention layers, respectively. The attention layer $\mathbf{att}(\boldsymbol{q}, \boldsymbol{k}, \boldsymbol{v})$ is defined as follows:

$$\mathbf{att}(\boldsymbol{q}, \boldsymbol{k}, \boldsymbol{v}) = \mathrm{FFN}\left(\mathrm{softmax}\left(\frac{\boldsymbol{q}\boldsymbol{k}^T}{\sqrt{d_k}}\right)\boldsymbol{v}\right)$$
$$\mathrm{FFN}(\boldsymbol{x}) = \boldsymbol{W}_2 \cdot \max(0, \boldsymbol{W}_1\boldsymbol{x} + \boldsymbol{b}_1) + \boldsymbol{b}_2 \quad (1)$$

where $\boldsymbol{q}$, $\boldsymbol{k}$, and $\boldsymbol{v}$ are the query, key, and value vectors, respectively. $d_k$ denotes the first dimension number of $\boldsymbol{k}$. FFN(·) represents the feed-forward sublayer that contains two linear transformations and an ReLU activation, where $\boldsymbol{W}_1$, $\boldsymbol{W}_2$, $\boldsymbol{b}_1$, and $\boldsymbol{b}_2$ are relevant weights and biases in two linear transformations, respectively.

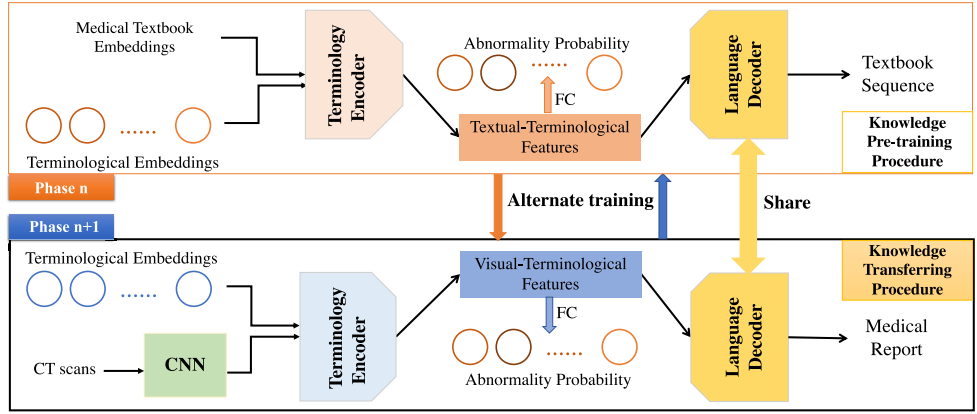

Fig. 2. Overview of our proposed Medical-VLBERT. The input images are first encoded as spatial features through a convolutional neural network (CNN) as the visual context. Then two separate terminology encoders are designed to associate the predefined terminology embeddings with the visual contexts and medical textbook embeddings, which produce the visual-terminological and textual-terminological features correspondingly. Besides, the alternate training strategy is exploited to minimize the discrepancies between these two terminology-related features. Based on the visual-terminological and textual-terminological features, a shared language decoder is employed to generate a report sequence and medical textbook sequence.

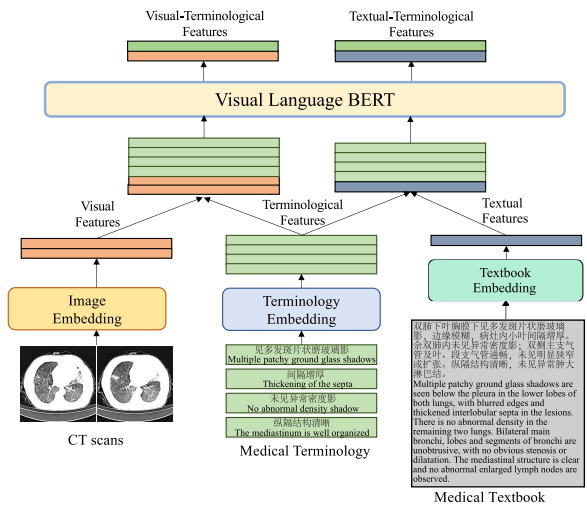

Fig. 3. Illustration of the terminology encoder. The terminology encoder is implemented in visual language BERT (VLBERT). Terminology embeddings are grouped with visual and textual embeddings, respectively, to form parallel corresponding inputs for VLBERT.

To jointly attend to the information from different domains in the attention layers, each terminology representation $\boldsymbol{m}_i$ needs to combine every visual feature pixel $\boldsymbol{v}_j$ to find mutual hidden relationships. Thus, we form a unified visual-terminological features $\mathcal{P} = \{\boldsymbol{p}_i\}_{i=1}^{N_m+N_v}$, which is defined as follows:

$$\mathcal{P} = \mathcal{M} \cup \mathcal{V} = \{\boldsymbol{p}_1, \ldots, \boldsymbol{p}_{N_m+N_v}\}$$
$$= \{\boldsymbol{m}_1, \ldots, \boldsymbol{m}_{N_m}, \boldsymbol{v}_1, \ldots, \boldsymbol{v}_{N_v}\} \quad (2)$$

where $\cup$ denotes the union operation. Every element in $\mathcal{P}$ is utilized as the input of attention layer, where self-attention and coattention are performed simultaneously

$$\boldsymbol{a}_{i*i} = \mathbf{att}(\boldsymbol{p}_i, \boldsymbol{p}_i, \boldsymbol{p}_i)$$
$$\boldsymbol{a}_{i*j} = \mathbf{att}(\boldsymbol{p}_i, \boldsymbol{p}_j, \boldsymbol{p}_i) \quad (3)$$

where $\boldsymbol{a}_{i*i}$ and $\boldsymbol{a}_{i*j}$ denote the visual-terminological vectors after self-attention and coattention, respectively. The processed visual-terminological features $\mathcal{A} = \{\boldsymbol{a}_i\}_{i=1}^{N_m+N_v}$ comprise the distributions of refined terminology-related features after the interaction between the visual features $\mathcal{V}$ and the original terminology features $\mathcal{M}$. Thus, they can be employed to determine which terminology is accurate to depict the signs in CT scans. The front elements of $\boldsymbol{a}_i \in \mathcal{A}$, $i \in [1, N_m]$, are extracted to output the updated terminology distribution.

The above formulation focuses more on the coherent relations between image regions and medical terminologies. Since the sentences in a medical report consist of a variety of terminologies, these medical terms also need to be highlighted in sentences, so as to aid sentence generation. To help identify the potential relationships between medical terminologies $\mathcal{M}$ and textbook embeddings $\mathcal{E}$, we also perform attention operations on these features. Similarly, we define a unified textual-terminological features $\mathcal{Q} = \{\boldsymbol{q}_i\}_{i=1}^{N_m+N_e}$

$$\mathcal{Q} = \mathcal{M} \cup \mathcal{E} = \{\boldsymbol{q}_1, \ldots, \boldsymbol{q}_{N_m+N_e}\}$$
$$= \{\boldsymbol{m}_1, \ldots, \boldsymbol{m}_{N_m}, \boldsymbol{e}_1, \ldots, \boldsymbol{e}_{N_e}\}. \quad (4)$$

Following that, the textual-terminological features $\mathcal{Q}$ are sent to the attention layer for further coherent relationship exploration

$$t_{i*i} = \mathbf{att}(q_i, q_i, q_i)$$
$$t_{i*j} = \mathbf{att}(q_i, q_j, q_i) \quad (5)$$

where $t_{i*i}$ and $t_{i*j}$ denote the textual-terminological vectors after self-attention and coattention, respectively. The obtained textual-terminological features $\mathcal{T} = \{t_i\}_{i=1}^{N_m+N_e}$ can be utilized for sentence generation since they contain the corresponding information between text embeddings $\mathcal{E}$ and medical terminologies $\mathcal{M}$. More specifically, they act like a grammar book that can provide references and norms for medical terminologies to generate professional sentences. For example, medical terminology "pulmonary vascularity" could be utilized to generate a simple sentence, such as "pulmonary vascularity is within normal limits," since $\mathcal{T}$ contains a connection between "spine" and "there are minimal degenerative changes of the spine." The front elements of $t_i \in \mathcal{T}$, $i \in [1, N_m]$, can also be exploited to scrutinize the acquired terminology distribution.

The acquired terminology distribution from textual-terminological features or visual-terminological features is used for abnormality classification. We treat the abnormality classification as a multilabel classification task and predict scores as follows:

$$x_i = \mathrm{sigmoid}(W_{\hat{m}}\hat{m}_i + b_{\hat{m}})$$
$$\mathcal{L}_{\mathrm{cls}} = \frac{1}{N_m} \sum_{i=1}^{N_m} \left[ y_i \log x_i + (1 - y_i) \log(1 - x_i) \right] \quad (6)$$

where $\hat{m}_i \in \mathcal{A}$ or $\mathcal{T}$, $i \in [1, N_m]$, $W_{\hat{m}}$, and $b_{\hat{m}}$ are parameters of a linear projection, respectively. This projection transforms terminology-related features into a distribution over all terminologies. $\mathcal{L}_{\mathrm{cls}}$ is the average binary cross-entropy loss, $x_i$ denotes the probability of abnormal medical term $\hat{m}_i$, $y_i$ is the ground-truth label, and $N_m$ is the number of abnormalities, which is the number of medical terminologies.

### C. Shared Language Decoder

We design a shared language decoder to perform alternate training and enable semantic alignment in linguistic and visual domains.

Given a set of medical textbook or report tokens $\mathcal{R} = \{r_i\}_{i=1}^{N_r}$, we adopt a standard language modeling objective [22] to maximize the likelihood in the following formulation:

$$P(r_1, r_2, \ldots, r_{N_r}; \Theta_{\mathcal{R}}) = \prod_{i=1}^{N_r} P\left(r_i | r_1, r_2, \ldots r_{i-1}; \Theta_{\{r_j\}_{j=1}^{i-1}}\right)$$
$$\mathcal{L}(\mathcal{R}) = -\sum_{i=1}^{N_r} \log P(r_i | r_1, \ldots, r_{i-1}; \Theta_{\mathcal{R}}) \quad (7)$$

where $P(r_i | r_1, r_2, \ldots, r_{i-1}; \Theta_{\{r_j\}_{j=1}^{i-1}})$ is the probability of next token conditioned in the history sequence and parameters $\Theta_{\{r_j\}_{j=1}^{i-1}}$, and $\mathcal{L}(\mathcal{R})$ denotes the corresponding loss in text generation.

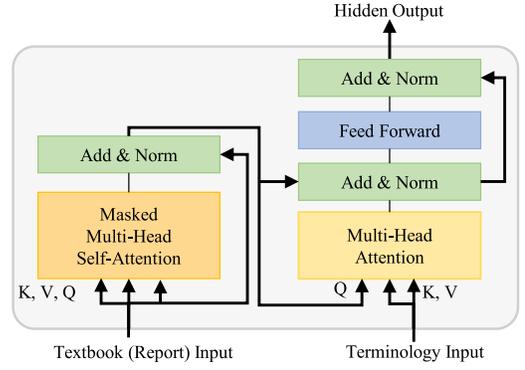

Fig. 4. Illustration of a **block** in language decoder. The terminology input denotes terminology-related features, i.e., visual-terminological features or textual-terminological features, from the terminology encoder. Text (report) input denotes word embeddings from textbook or report.

Based on the history sequence $\{r_1, \ldots, r_{i-1}\}$, our shared language decoder applies a multihead attention operation over the medical terminology-related features, i.e., visual-terminological features, or textual-terminological features, followed by positionwise feed-forward networks to compute a probability distribution $P(i)$ over tokens in the vocabulary

$$h_0 = w\mathrm{E}_w + p\mathrm{E}_p$$
$$h_l = \mathbf{block}(h_{l-1}, \hat{m}) \quad \forall l \in [1, N]$$
$$P(i) = \mathrm{softmax}(h_N \mathrm{E}_w^T) \quad (8)$$

where $w$ is the vocabulary index vector, $p$ is the position index vector, $\mathrm{E}_w$ is the word embedding matrix, $\mathrm{E}_p$ is the position embedding matrix the index vector of position, and **block** is the architecture of transformer decoder block, which is depicted in Fig. 4.

### D. Alternate Training Strategy

In terms of data distribution and input source information (language or vision), there exists a large semantic deviation between the collected medical textbooks and image report datasets. Unlike recent language representation models [5], [22] that are trained in an unsupervised manner, we develop a strategy in Algorithm 1 to alleviate data bias and narrow the gap between language and vision domains.

---

**Algorithm 1** Alternate Training Strategy

1: Initialize the shared Language Decoder ($\mathcal{D}$);
2: **repeat**
3:   Load $\mathcal{D}$ and execute the pre-training procedure;
4:   Optimize $\mathcal{D}$;
5:   Stop the pre-training procedure;
6:   Load $\mathcal{D}$ and execute the transferring procedure;
7:   Optimize $\mathcal{D}$.
8:   Stop the transferring procedure;
9: **until** Convergence

---

In each epoch, we train the shared language decoder with terminology encoder in an end-to-end manner to optimize the objective function discussed in (9). Specifically, the knowledge pretraining procedure is first executed for the first epoch to optimize the decoder. In the next epoch, the pretraining procedure is halted, and the transferring procedure is started to fine-tune the trained decoder. Finally, the trained decoder is reloaded in the pretraining procedure, and the above operation is repeated until convergence. With the alternate training strategy, these two procedures can exchange knowledge and adapt mutually.

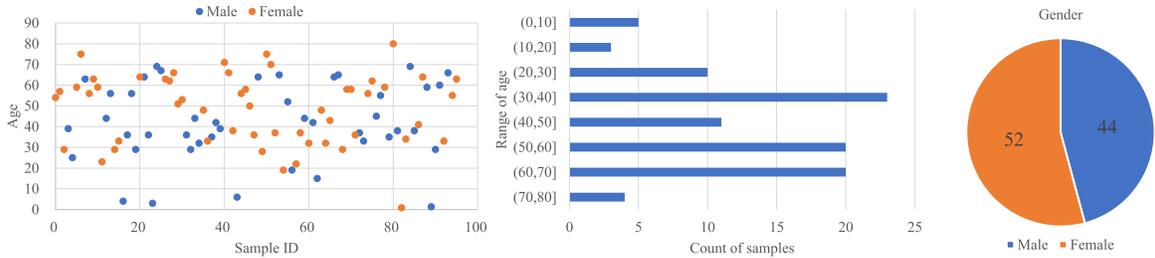

Fig. 5. Age and gender distribution of the COVID-19 patients in our COVID-19 CT dataset. The range of age is from 10-month old to 80-year old.

### E. Optimization

Training samples consist of tuples $(y, \mathcal{T})$ or triplets $(I, y, \mathcal{T})$ in pretraining or transferring procedure, respectively, where $I$ is an image, $y$ denotes the ground-truth abnormalities, and $\mathcal{T}$ is the original textbook sequence or ground-truth report. Formally, we define a multitask loss as follows:

$$\mathcal{L} = \lambda \mathcal{L}_{\text{cls}} + \mathcal{L}_{\mathcal{T}} \qquad (9)$$

where $\mathcal{L}_{\text{cls}}$ is the terminology multilabel classification loss defined in (6), $\mathcal{L}_{\mathcal{T}}$ is token-level cross entropy loss defined in (7), and $\lambda = 1$ is a balance weight. The encoder–decoder architecture is jointly trained to minimize $\mathcal{L}$.

## IV. EXPERIMENTS

### A. Dataset

Two medical image datasets are employed to evaluate the performance of our proposed Medical-VLBERT.

*1) COVID-19 CT:* The COVID-19 CT dataset is a collection of the chest CT images with Chinese reports for the COVID-19 checking, consisting of 368 medical reports and 1104 CT images from 96 patients. We count the details of these patients, and the distribution of ages, counts of samples in age range, and the gender percentages are shown in Fig. 5. In the raw data, each report accompanies a set of CT images. We filter ten CT images characterizing distinct features for each report and build our own COVID-19 CT dataset for report generation.

*2) CX-CHR:* It is a **large-scale** collection of CXR images with Chinese reports for health checking, consisting of 45 598 images of 35 609 patients and 28 299 medical image-report pairs. In addition, CX-CHR includes 12 million external medical textbooks from a Chinese medical website,[3] which contains symptoms, manifestations, laboratory tests, and other information on various diseases in thoracic surgery.

[3]http://www.fh21.com.cn

TABLE I
ARCHITECTURES OF COMPARED METHODS. CE IS SHORT FOR
CROSS-ENTROPY. RL DENOTES THE DECODER OPTIMIZED
BY REINFORCEMENT LEARNING. TEMPLATE DENOTES
THE DECODER GENERATING SENTENCES
FROM A TEMPLATE DATABASE

| Models | Encoder | Decoder | Strategy |
|---|---|---|---|
| CoAtt [2] | CNN | RNN | CE |
| HRGR-Agent [1] | CNN | RNN+Template | CE+RL |
| KERP [19] | CNN+Attention | Template | CE |
| Vision-BERT [5] | CNN | Pre-trained Decoder | CE |
| DenseNet+GPT-2 | Transformer Encoder | Pre-trained Decoder | CE+Alternate |
| **Medical-VLBERT** | VLBERT | Pre-trained Decoder | CE+Alternate |

### B. Implementation Details

We implement our method on Pytorch Framework and conduct the training on two GPUs. **DenseNet-121** [35] is adopted as the backbone, which takes a $224 \times 224$ CT image as input to achieve (7, 7, 1024) feature maps from the last convolution layer. The dimension of all liner projection layers is set to 512. To balance language model perplexity with model size and computational requirements, we set the embedding and the hidden size of both **VLBERT** and language decoder to 512. The numbers of hidden layers and attention heads of the terminology encoder are 2 and 8, respectively. Due to the better performance achieved by concatenation than summation in practice, we adopt concatenation as the feature fusion approach. To minimize the loss function, we employ the same ADAM optimizer in training with a batch size of 32. The backbone is trained with a learning rate of $10^{-6}$, and the two knowledge procedures are trained with a learning rate of $5 \times 10^{-5}$ for 30 epochs. In the beginning, we train the model on the large-scale CX-CHR dataset by the alternate training strategy. Then, we fine-tune the model on our COVID-19 CT dataset with the alternate training strategy as well. Due to the scale of the COVID-19 CT dataset, we enhance the transferring procedure, which means that we execute the pretraining procedure once and the transferring procedure multiple times at one epoch.

### C. Evaluation Metrics

The evaluation of our experiments consists of objective automatic evaluation and subjective human evaluation. To provide fair and clear evaluation results of our model, different metrics are employed for such two types of evaluations.

*1) Objective Evaluation:* The evaluation metrics for objective automatic evaluation contain BLEU (unigram to 4-gram), ROUGE-L, and CIDEr-D. BLEU [36] is a classic algorithm for automatic evaluation of machine translation. It measures the consensus between the machines' output and the human being's output and is, thus, applied to test the accuracy of generated reports. ROUGE-L [37] is another most-used metric in machine translation. On the basis of the longest common subsequence statistics, ROUGE-L computes the correlation between a candidate translation and other references. Thus, it is suitable to measure the correspondence between the produced sentences and the original expressions. CIDEr-D [38] is a specialized metric for image captioning evaluation that measures the similarities between image descriptions and reference sentences. Therefore, it is used in our evaluation of the text generation based on the COVID-19 CT scans.

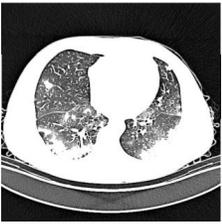

Fig. 6. Illustrations of the reports generated by Medical-VLBERT and DenseNet+GPT-2. The underlined sentences are the descriptions of lesions that match the original ones in the ground truth. The selected CT scans are from the test dataset.

*2) Subjective Evaluation:* To give more convincing performances of our presented model, we further conduct the subjective human evaluation by radiologists following HRGR-Agent [1]. We randomly select 16 CT scans of the COVID-19 patients and the corresponding reports generated by Medical-VLBERT, DenseNet+GPT-2, and professional radiologists. Here, Medical-VLBERT is our proposed model, and DenseNet+GPT-2 is the model using GPT-2 [39] as terminology encoder and using the alternate training strategy. Three native Chinese radiologists are invited to view three reports and provide rankings to sort out, which is the best produced report. Specifically, the best report is categorized as 1, the second best one is categorized as 2, and the worst one is classified as 3. Then, we count the sum of categories of 1, 2, and 3 for three reports, respectively.

### D. Quantitative Results

*1) CX-CHR:* As shown in Table I, we compare our proposed model with other medical report generation methods. Table II shows the performance of these models on the CX-CHR dataset. The results across automatic evaluation metrics consistently indicate that, by exploiting the medical-specific knowledge recorded in large-scale unlabeled medical textbooks, the proposed Medical-VLBERT achieves superior performance than all the state-of-the-art techniques [1], [19].

Furthermore, to verify the capability of the language-to-vision transfer, Medical-VLBERT is compared with Vision-BERT [5] by combining our terminology encoder and the BERT model. As shown in the fourth row of Table II, out model outperforms other models significantly: it improves KERP on CIDEr-D by 37.0%, ROUGE-L by 3.7%, BLEU-1 by 2.7%, BLEU-2 by 3.9%, BLEU-3 by 3.8%, and BLEU-4 by 6.1%. Comparing with DenseNet + GPT-2, as shown in the fifth row in Table II, our terminology encoder performs better.

*2) COVID-19 CT:* After training on the CX-CHR dataset, we use the pretrained model and fine-tune on the COVID-19 CT dataset. In order to compare with our Medical-VLBERT, we adopt DenseNet+GPT-2 as the baseline, where DenseNet [35] encodes CT image, transformer [34] encoder acts as the terminology encoder, and GPT-2 acts as the language decoder. Due to the small scale of the COVID-19 data, we conduct different experiments with different alternate training strategies. As the results shown in Table III, a proper strategy is very important for our Medical-VLBERT, and it could boost around 100% in CIDEr-D score. However, for baseline, more transferring procedures in one epoch could only improve the BLEU score and even decrease CIDEr-D and ROUGE-L scores. As default, we will execute a one-time pretraining procedure and a three-time transferring procedure in one epoch in other experiments.

8 IEEE TRANSACTIONS ON NEURAL NETWORKS AND LEARNING SYSTEMS

This article has been accepted for inclusion in a future issue of this journal. Content is final as presented, with the exception of pagination.TABLE II
PERFORMANCE COMPARISON ON THE LARGE-SCALE CX-CHR DATASET

| Models | CIDEr-D | ROUGE-L | BLEU@1 | BLEU@2 | BLEU@3 | BLEU@4 |
|---|---|---|---|---|---|---|
| CoAtt [2] | 273.5 | 64.5 | 64.7 | 57.5 | 52.5 | 48.7 |
| HRGR-Agent [1] | 289.5 | 61.2 | 67.3 | 58.7 | 53.0 | 48.6 |
| KERP [19] | 285.0 | 61.8 | 67.3 | 58.8 | 53.2 | 47.3 |
| Vision-BERT [5] | 302.4 | 63.7 | 68.6 | 60.1 | 54.1 | 50.3 |
| DenseNet + GPT-2 | 317.7 | 64.6 | 68.8 | 60.7 | 55.7 | 52.1 |
| **Medical-VLBERT (Ours)** | **322.0** | **65.5** | **70.0** | **62.7** | **57.0** | **53.4** |

TABLE III
PERFORMANCE COMPARISON ON DIFFERENT ALTERNATE TRAINING STRATEGIES. ($m$, $n$) MEANS THAT EXECUTE $m$ TIMES PRETRAINING PROCEDURE AND $n$ TIMES TRANSFERRING PROCEDURE AT ONE EPOCH

| Models | Strategy | CIDEr-D | ROUGE-L | BLEU@1 | BLEU@2 | BLEU@3 | BLEU@4 |
|---|---|---|---|---|---|---|---|
| DenseNet + GPT-2 | (1,1) | 23.2 | 58.9 | 32.1 | 31.0 | 30.3 | 29.7 |
|  | (1,3) | 23.1 | 51.4 | 55.3 | 47.6 | 42.3 | 39.0 |
|  | (1,4) | 17.7 | 51.5 | 51.5 | 44.4 | 40.4 | 37.7 |
|  | (1,5) | 15.6 | 52.9 | 56.6 | 48.3 | 43.4 | 40.3 |
| Medical-VLBERT (Ours) | (1,1) | 20.7 | 51.6 | 55.6 | 47.3 | 42.8 | 39.6 |
|  | (1,3) | 115.8 | 59.1 | 61.8 | 55.1 | 51.0 | 48.4 |
|  | (1,4) | **122.8** | 59.6 | 61.6 | 55.6 | 51.9 | 49.4 |
|  | (1,5) | 117.4 | **60.3** | **63.4** | **57.2** | **53.6** | **51.3** |

TABLE IV
PERFORMANCE COMPARISON ON OUR COVID-19 CT DATASET. MEDICAL-VLBERT@$N$ IS SHORT FOR MEDICAL-VLBERT WITH $N$ HIDDEN LAYERS

| Models | CIDEr-D | ROUGE-L | BLEU@1 | BLEU@2 | BLEU@3 | BLEU@4 |
|---|---|---|---|---|---|---|
| CoAtt [2] | 25.5 | 57.4 | 60.8 | 53.5 | 49.4 | 46.8 |
| Vision-BERT [5] | 29.6 | 57.8 | 57.9 | 51.4 | 47.5 | 44.9 |
| DenseNet + GPT-2 | 23.1 | 51.4 | 55.3 | 47.6 | 42.3 | 39.0 |
| VGG + GPT-2 | 31.5 | 51.3 | 55.8 | 48.0 | 42.6 | 39.2 |
| Medical-VLBERT@6 | 76.6 | 57.2 | 59.3 | 52.5 | 48.5 | 25.9 |
| Medical-VLBERT@4 | 97.6 | 57.5 | 59.4 | 52.9 | 49.0 | 46.5 |
| Medical-VLBERT@2 | **115.8** | **59.1** | **61.8** | **55.1** | **51.1** | **48.4** |

TABLE V
RESULTS OF TAG CLASSIFICATION

| Models | Precision(%) | Recall(%) | F1(%) |
|---|---|---|---|
| CoAtt [2] | 16.2 | **66.7** | 19.9 |
| Densenet [35] | 21.4 | 21.6 | 20.9 |
| Resnet [7] | 42.8 | 26.8 | 28.9 |
| VGG [40] | 23.4 | 20.9 | 20.0 |
| Medical-VLBERT | **64.7** | 49.0 | **52.3** |

Then, we explore the impact of the number of hidden layers of Medical-VLBERT. As shown in Table IV, we adopt CoAtt [2], Vision-BERT [5], VGG+GPT-2, and DenseNet+GPT-2 as the baselines. Two hidden layers perform the best results in all metrics. As the number of hidden layers increases, the parameters increase as well. Though the expression ability of the model is larger, it will converge slowly and require more independent data.

As mentioned before, there is also an abnormality classification module in our model. As shown in Table V, we adopt DenseNet, Resnet, and VGG as the baselines to do the multilabel classification directly. We can see our model outperforms the baselines, which means that the terminology encoder can reinforce the feature extraction of abnormalities.

*E. Visualization Results*

The medical reports generated by our Medical-VLBERT, DenseNet+GPT-2, and professional radiologists (ground truth) are shown in Fig. 6. Note that Medical-VLBERT produces paragraphs of CT scans which is more similar to that of professional radiologists than DenseNet+GPT-2. More specifically, Medical-VLBERT has the advantage of generating more accurate medical terminologies as well as more concise descriptions. By comparison, DenseNet+GPT-2 seemingly learns to provide a long-winded report template but has fewer matched medical terminologies.

The visualized results on the CX-CHR dataset are provided in Fig. 7. The generated reports demonstrate the significant alignment with ground-truth reports. Furthermore, Fig. 7(b) illustrates how knowledge of textbooks influences the generated reports.

Fig. 8 provides the visualization of activation mapping of our model on the COVID-19 CT dataset. The maps demonstrate the alignment with the lesion areas. During the inference, radiologists can judge the correctness of the generated report by checking the activation maps.

The attention maps of three textbooks are presented in Fig. 9. The dark patches demonstrate that the knowledge pretraining procedure helps learn medical-specific and terminology-related knowledge contained in the medical textbooks. The knowledge transferring procedure helps transfer the knowledge and bridge the gap between language and vision for report generation. Therefore, our proposed

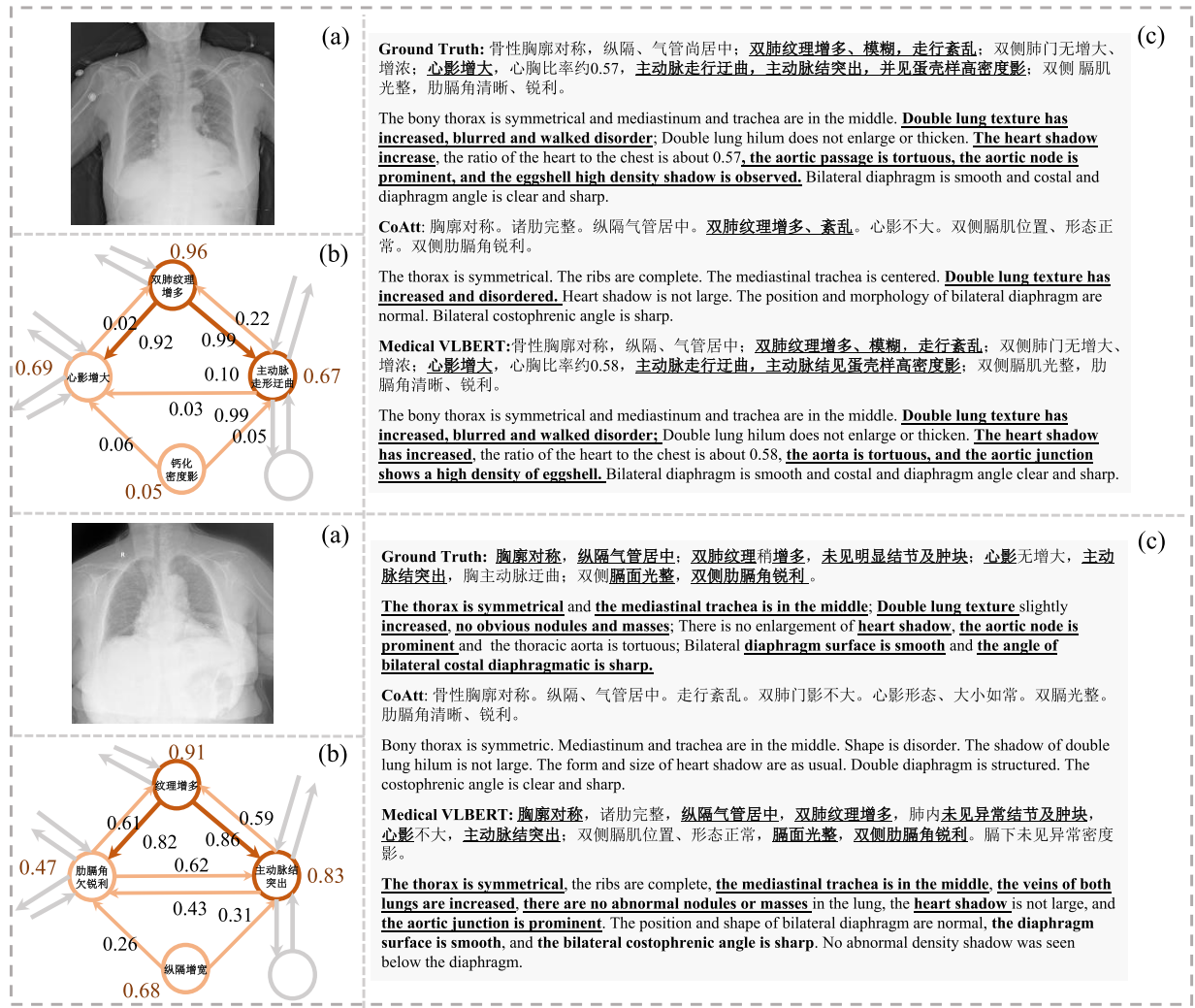

Fig. 7. (a) Visualization of the results generated by CoAtt [2] and Medical-VLBERT on the CX-CHR dataset. (b) Orange digits represent classification scores, and black digits denote attention weights. (c) Underlined texts represent the alignment between generated and ground-truth reports.

Medical-VLBERT can provide a more satisfactory report for the COVID-19 diagnosis in practice.

### F. Subjective Evaluation by Radiologists

We make a summary of the subjective evaluation by radiologists in Fig. 10. It is evident that report obtained by our proposed Medical-VLBERT receives the second most votes as the best report (Category 1), merely fewer votes than that of humans, i.e., ground-truth reports, by 6. Medical-VLBERT's report also has better evaluation than DenseNet+GPT-2's one since the report generated by DenseNet+GPT-2 is mostly voted as the worst report (Category 2). The results in Fig. 10 can well demonstrate that Medical-VLBERT has an excellent performance in report generation and produce medical texts more similar to that of humans.

### G. Ablation Studies

To verify the most important components of our proposed Medical-VLBERT model, we conduct the ablation studies via replacing or removing specific components within our model. All results of ablation studies are displayed in Table VI.

*1) External Knowledge:* External medical knowledge plays a significant role in our model training. The introduction of external knowledge by pretraining the language decoder in the medical domain corpus further boosts the model's performance. According to the results of the first and third row in Table VI, BLEU increases significantly, but CIDEr-D and ROUGE-L decrease. Due to the small scale of the COVID-19 CT dataset, the model tends to overfit without external knowledge. As a result, the model is prone to produce more words from the training set, leading to high ROUGE-L. However, if we introduce the EK into the model, it would learn more general medical knowledge and generate fewer unrelated words, resulting in a high BLEU. Moreover, CIDEr-D assigns low weight to high-frequency phrases, but there are many keywords with high frequency in the dataset for medical report generation. According to the results of the second and fourth rows in the Tabel VI, the addition of the EK only increases CIDEr-D. However, as presented in the last four rows in Table VI, the addition of the EK to the model with the ATS boosts the performances on all the metrics, which demonstrates that the strategy of introducing external knowledge is useful.

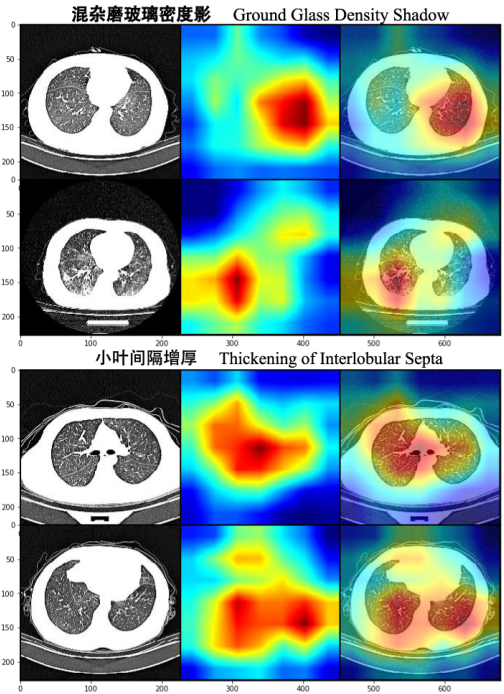

Fig. 8. Visualization of the activation mapping generated by Medical-VLBERT on the COVID-19 CT dataset. The patterns of *"Ground Glass Density Shadow"* and *"Thickening of Interlobular Septa"* denote the existence of some ground glass shadows and lots of fine and linear white shadows, respectively.

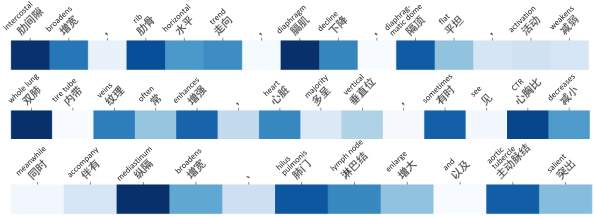

Fig. 9. Attention maps of three textbooks indicate what knowledge the pretraining procedure has acquired.

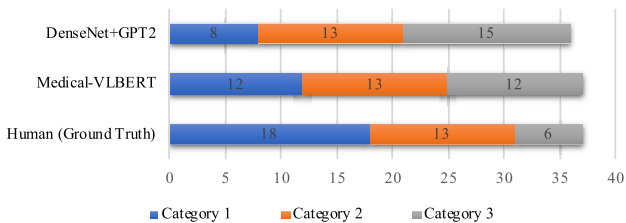

Fig. 10. Subjective evaluation by three radiologists on the reports generated by Medical-VLBERT, DenseNet+GPT-2, and humans, where the best report is categorized as 1, the second best as 2, and the worst as 3.

*2) Alternate Training:* We design an experiment to validate the effectiveness of the alternate training strategy on the condition that external knowledge exists. The Medical-VLBERT is first trained on collected medical textbooks and then trained on CT images for 30 epochs, which means that knowledge pretraining and transferring procedures are separately executed. As expected, experimental results in the last four rows in Table VI show that our model's performance is weakened without alternate training. Due to the lack of efficient knowledge transferring, the shared language decoder is easily affected by data bias and domain misalignment between language and vision. Thus, the model is unable to effectively make use of the linguistic knowledge in medical text and, as a result, generate unsatisfactory paragraphs. Therefore, an alternate training strategy enables two procedures to adapt mutually, and the shared language decoder of the model receives better optimization.

TABLE VI
PERFORMANCE CONTRIBUTION OF EACH COMPONENT IN MEDICAL-VLBERT ON THE COVID-19 CT. ATS, TLS, AND EK ARE SHORT FOR ALTERNATE TRAINING STRATEGY, TRANSFER LEARNING STRATEGY, AND EXTERNAL KNOWLEDGE

| Components | | | Evaluation Metrics | | | |
|---|---|---|---|---|---|---|
| ATS | TLS | EK | CIDEr-D | ROUGE-L | BLEU@-1 | BLEU@-4 |
| - | - | - | 110.2 | **60.6** | 47.9 | 39.9 |
| - | ✓ | - | $28.5^{\downarrow 81.7}$ | $56.3^{\downarrow 4.3}$ | $57.5^{\uparrow 9.6}$ | $44.2^{\uparrow 4.3}$ |
| - | - | ✓ | $40.6^{\downarrow 69.6}$ | $55.7^{\downarrow 4.9}$ | $58.7^{\uparrow 10.8}$ | $45.1^{\uparrow 5.2}$ |
| - | ✓ | ✓ | $38.3^{\downarrow 71.9}$ | $55.9^{\downarrow 4.7}$ | $56.0^{\uparrow 8.1}$ | $42.6^{\uparrow 2.7}$ |
| ✓ | - | ✓ | $95.7^{\downarrow 14.5}$ | $59.0^{\downarrow 1.6}$ | $59.0^{\uparrow 11.1}$ | $46.6^{\uparrow 6.7}$ |
| ✓ | ✓ | ✓ | **$115.8^{\uparrow 5.6}$** | $59.1^{\downarrow 1.5}$ | **$61.8^{\uparrow 13.9}$** | **$48.4^{\uparrow 8.5}$** |

*3) Transfer Learning:* We adopt a transfer learning strategy to alleviate the shortage of the available COVID-19 data. We conduct experiments to verify whether the model has a better performance after transferring from the large-scale CX-CHR dataset. As shown in Table VI, the results of the first and second rows on CIDEr-D and BLEU prove that the model is able to exploit the knowledge from the CX-CHR dataset and improve the performance on the COVID-19 dataset. As the results of the third and fourth rows shown in Table VI, the TLS cannot provide improvement to the model without the ATS. Nevertheless, as the fifth and sixth rows displayed in Table VI, the addition of the TLS gives better results with the ATS, which implies that the ATS could enhance the advantage of TLS.

## V. CONCLUSION

In this work, we proposed the Medical-VLBERT model for automatic medical report generation on the COVID-19 CT scans. Medical-VLBERT further adopts an alternate learning strategy that can use the transferred medical textual knowledge to identify the abnormality in CT images and generate the medical report automatically based on the detected lesion regions with acquired knowledge. Besides, we built a COVID-19 CT dataset of 368 medical findings in Chinese and 1104 chest CT scans from The First Affiliated Hospital of Jinan University and The Fifth Affiliated Hospital of Sun Yat-sen University. To solve the COVID-19 data shortage problem, we employed a transfer learning fashion in which the model is first trained on the large-scale CX-CHR dataset in Chinese and then fine-tuned on the COVID-19 CT dataset. Results and analysis in our experiments proved that our model has state-of-the-art performances on terminology prediction and report generation on the Chinese COVID-19 CT dataset and the Chinese CX-CHR dataset. Furthermore, in clinical practice, our model could help ease the radiologists' burdens via aiding their diagnostic workflow, including analyzing the lesions in images and writing corresponding medical reports for diagnosed patients. In the future, we will pay more attention to the diagnostic phase information on a patient's timelines and focus on the generation of more integrated reports.

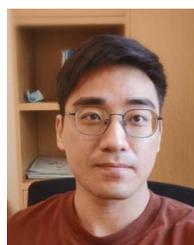

**Guangyi Liu** (Graduate Student Member, IEEE) received the B.E. degree in automation engineering from the University of Electronic Science and Technology of China, Chengdu, China, in 2018. He is currently pursuing the Ph.D. degree with the Deep Bit Lab, Future Network of Intelligence Institute (FNii), The Chinese University of Hong Kong, Shenzhen, China.

His research interests include natural language generation and text variational autoencoder.

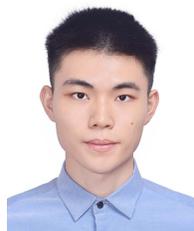

**Yinghong Liao** (Graduate Student Member, IEEE) received the B.E. degree in software engineering from the School of Data and Computer Science, Sun Yat-sen University, Guangzhou, China, in 2019. He is currently pursuing the Ph.D. degree with the Deep Bit Lab, Future Network of Intelligence Institute (FNii), The Chinese University of Hong Kong, Shenzhen, China.

His research interests include domain adaptation, metalearning, and multimodal learning.

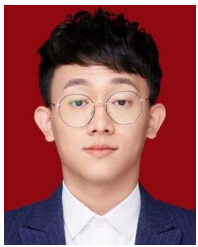

**Fuyu Wang** received the B.E. degree in software engineering from the School of Computer Science and Engineering, Sun Yat-sen University, Guangzhou, China, in 2018, where he is currently pursuing the Ph.D. degree in computer science.

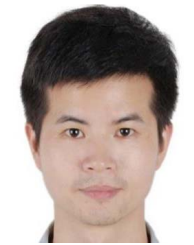

**Bin Zhang** received the M.M. degree from Southern Medical University, Guangzhou, China, in 2017, and the M.D. degree from Jinan University, Guangzhou, in 2021.

His research interests include radiomics and deep learning.

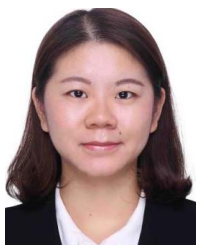

**Lu Zhang** received the M.M. degree from Southern Medical University, Guangzhou, China, in 2019. She is currently pursuing the M.D. degree with Jinan University, Guangzhou.

Her research interests include radiomics and deep learning.

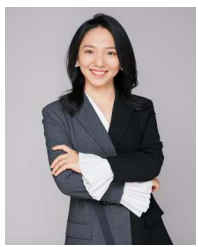

**Xiaodan Liang** received the Ph.D. degree from Sun Yat-sen University, Guangzhou, China, in 2016, advised by Liang Lin.

She was a Post-Doctoral Researcher with the Machine Learning Department, Carnegie Mellon University, Pittsburgh, PA, USA, working with Prof. Eric Xing, from 2016 to 2018. She is currently an Associate Professor with Sun Yat-sen University. She has published several cutting-edge projects on human-related analysis, including human parsing, pedestrian detection and instance segmentation, 2-D/3-D human pose estimation, and activity recognition.

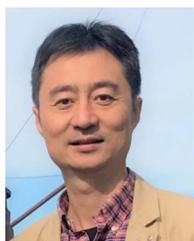

**Xiang Wan** received the B.A. degree in information system from Renmin University, Beijing, China, in 1994, and the M.A. and Ph.D. degrees in computing science from the University of Alberta, Edmonton, AB, Canada, in 2002 and 2006, respectively.

He was a Research Assistant Professor with Hong Kong Baptist University, Hong Kong, from 2012 to 2018. He has been a Senior Research Scientist with the Shenzhen Research Institute of Big Data, Shenzhen, China, since 2018. He is currently with Guangdong Provincial Key Laboratory of Big Data Computing, The Chinese University of Hong Kong, Shenzhen. He has been mainly working on meta-analysis and statistical learning, particularly in the field of large-scale genomic data analysis. He has published more than 50 articles in many top-tier journals and conferences.

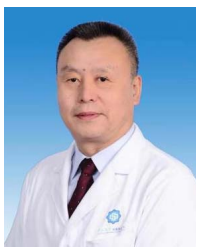

**Shaolin Li** received the B.M. degree from the Department of Medicine, Air Force Medical University, Xi'an, China, in 1986, and the M.M. and M.D. degrees in imaging and nuclear medicine from First Military Medical University, Guangzhou, China, in 1992 and 2007, respectively.

He is currently a Professor and the Chair of the Department of Imaging Medicine and the Department of Radiology, The Fifth Affiliated Hospital of Sun Yat-sen University, Zhuhai, China.

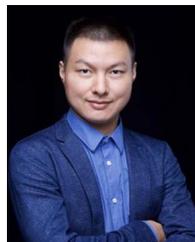

**Zhen Li** received the bachelor's degree in automation and the master's degree in communication and information systems from Sun Yat-sen University, Guangzhou, China, in 2011 and 2014, respectively, and the Ph.D. degree in computer science from The University of Hong Kong, Hong Kong in 2018.

He worked as a Visiting Scholar with The University of Chicago, Chicago, IL, USA, in 2018, and a Visiting Student with the Toyota Technological Institute at Chicago (TTIC), Chicago, in 2016. He is currently serving as an Assistant Professor with the School of Science and Engineering and a Research Scientist with Shenzhen Research Institute of Big data, The Chinese University of Hong Kong, Shenzhen (CUHKSZ), China. He has published many papers in top-tier conferences and journals. His research interests include medical imaging and medical big data, AI interdisciplinary research, and computer vision.

Dr. Li was the Core Team Member for the Champion of the 12th Critical Assessment of Protein Structure Prediction (CASP12), with the published paper receiving the PLOS CB 2018 Breakthrough and Innovation Awards and being the Web-of-Science Highly Cited Paper.

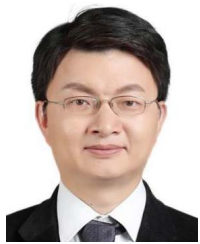

**Shuixing Zhang** received the B.S., M.S., and Ph.D. degrees from the Department of Radiology, First Military Medical University, Guangzhou, China, in 1993, 2004, and 2007, respectively.

He is currently a Professor and the Chair of the Department of Radiology, The First Affiliated Hospital of Jinan University, Guangzhou.

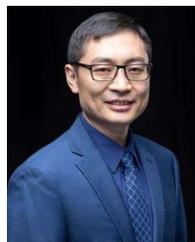

**Shuguang Cui** (Fellow, IEEE) received the Ph.D. degree in electrical engineering from Stanford University, Stanford, CA, USA, in 2005.

He has an Assistant Professor, an Associate Professor, a Full Professor, and a Chair Professor of electrical and computer engineering with The University of Arizona, Tucson, AZ, USA, Texas A&M University, College Station, TX, USA, the University of California at Davis (UC Davis), Davis, CA, USA, and The Chinese University of Hong Kong, Shenzhen (CUHKSZ), China, respectively. He has also been the Executive Dean of the School of Science and Engineering and the Executive Vice Director of the Shenzhen Research Institute of Big Data, CUHKSZ. His current research interests focus on data-driven large-scale system control and resource management, large dataset analysis, the IoT system design, energy-harvesting-based communication system design, and cognitive network optimization.

Dr. Cui was a member of the IEEE ComSoc Emerging Technology Committee. He was an elected member of the IEEE Signal Processing Society's SPCOM Technical Committee from 2009 to 2014 and the elected Chair of the IEEE ComSoc Wireless Technical Committee from 2017 to 2018. He is also a member of the Steering Committee of the IEEE TRANSACTIONS ON BIG DATA and the Chair of the Steering Committee of the IEEE TRANSACTIONS ON COGNITIVE COMMUNICATIONS AND NETWORKING. He has won the IEEE ICC Best Paper Award, the ICIP Best Paper Finalist, the First Class Prize in Natural Science from the Chinese Institute of Electronics, and the First Class Prize in Technology Invention from the China Institute of Communications all in 2020. He was selected as the Thomson Reuters Highly Cited Researcher and listed in the Worlds' Most Influential Scientific Minds by ScienceWatch in 2014. He was a recipient of the IEEE Signal Processing Society 2012 Best Paper Award. He has served as the general co-chair and the TPC co-chair of many IEEE conferences. He has been serving as the Area Editor for the *IEEE Signal Processing Magazine* and an Associate Editor for the IEEE TRANSACTIONS ON BIG DATA, IEEE TRANSACTIONS ON SIGNAL PROCESSING, IEEE JSAC Series on Green Communications and Networking, and IEEE TRANSACTIONS ON WIRELESS COMMUNICATIONS. He was elected as an IEEE ComSoc Distinguished Lecturer in 2014 and an IEEE VT Society Distinguished Lecturer in 2019.